\documentclass[10pt,twocolumn,letterpaper]{article}

\usepackage{iccv}
\usepackage{url}
\usepackage{times}
\usepackage{epsfig}
\usepackage{graphicx}
\usepackage{amsmath}
\usepackage{amssymb}
\usepackage[titletoc]{appendix}
\usepackage{color}
\usepackage{ulem}
\usepackage{ifthen}
\usepackage{algorithm}
\usepackage[noend]{algpseudocode}
\usepackage{array,multirow}
\usepackage{adjustbox}
\usepackage{hhline}
\usepackage[justification=raggedright]{caption}	
\definecolor{gray}{rgb}{0.5,0.5,0.5} 
\definecolor{green}{rgb}{0, 0.4, 0} 
\definecolor{orange}{rgb}{1, 0.5, 0} 	
\definecolor{mahogany}{rgb}{0.75, 0.25, 0.0}
\definecolor{purple}{rgb}{0.6, 0, 0.6}
\definecolor{purple}{rgb}{0.6, 0, 0.6}
\definecolor{darkgreen}{rgb}{0, 0.4, 0} 
\definecolor{frenchblue}{rgb}{0.0, 0.45, 0.73}
\newboolean{revising}
\setboolean{revising}{true}
\ifthenelse{\boolean{revising}}
{
	\newcommand{\ignore}[1]{}

} {
	\newcommand{\ignore}[1]{}

}

\usepackage[pagebackref=true,breaklinks=true,letterpaper=true,colorlinks,bookmarks=false]{hyperref}

\newcommand*{\affaddr}[1]{#1} 
\newcommand*{\affmark}[1][*]{\normalsize\textsuperscript{#1}}
\newcommand*{\email}[1]{\normalsize\texttt{#1}}

\iccvfinalcopy 

\def\iccvPaperID{694} 

\ificcvfinal\fi
\begin{document}

\title{No More Discrimination: Cross City Adaptation of Road Scene Segmenters}

\author{
\normalsize
Yi-Hsin Chen\affmark[1],
Wei-Yu Chen\affmark[3,4],
Yu-Ting Chen\affmark[1]\thanks{indicates equal contribution}  ,
Bo-Cheng Tsai\affmark[2]\footnotemark[1]  ,
Yu-Chiang Frank Wang\affmark[4],
Min Sun\affmark[1] \\ \\
\affaddr{\normalsize Department of \{\affmark[1]Electrical Engineering,\affmark[2]Communication Engineering\}, National Tsing Hua University, Taiwan} \\
\affaddr{\affmark[3]Department of Electrical Engineering,
National Taiwan University, Taiwan} \\
\affaddr{\affmark[4]Research Center for Information Technology Innovation,
Academia Sinica, Taiwan} \\
\email{\{yhethanchen, wyharveychen, yuting2401, vigorous0503\}@gmail.com} \\
\email{, ycwang@citi.sinica.edu.tw}
\email{, sunmin@ee.nthu.edu.tw}
}

\maketitle

\begin{abstract}
Despite the recent success of deep-learning based semantic segmentation, deploying a pre-trained road scene segmenter to a city whose images are not presented in the training set would not achieve satisfactory performance due to dataset biases. Instead of collecting a large number of annotated images of each city of interest to train or refine the segmenter, we propose an unsupervised learning approach to adapt road scene segmenters across different cities. By utilizing Google Street View and its time-machine feature, we can collect unannotated images for each road scene at different times, so that the associated static-object priors can be extracted accordingly. By advancing a joint global and class-specific domain adversarial learning framework, adaptation of pre-trained segmenters to that city can be achieved without the need of any user annotation or interaction. We show that our method improves the performance of semantic segmentation in multiple cities across continents, while it performs favorably against state-of-the-art approaches requiring annotated training data.
\end{abstract}
\def \Ltotal {\mathcal{L}_{total}}
\def \Ltask {\mathcal{L}_{task}}
\def \Lga {\mathcal{L}_{G}}
\def \Lca {\mathcal{L}_{class}}
\def \lmga {\lambda_{G}}
\def \lmca {\lambda_{class}}

\def \Mf {M_F}
\def \My {M_Y}
\def \Mga {M_{G}}
\def \Mca {M_{class}^c}

\def \thf {\theta_F}
\def \thy {\theta_Y}
\def \thga {\theta_{G}}
\def \thca {\theta_{class}^{c}}

\def \S  {\mathcal{S}} 
\def \T  {\mathcal{T}} 
\def \Is {I_\mathcal{S}} 
\def \It {I_\mathcal{T}} 
\def \is {i_\mathcal{S}} 
\def \it {i_\mathcal{T}} 
\def \ns {n_\mathcal{S}} 
\def \nt {n_\mathcal{T}} 

\def \fx {\Mf(x,\thf)} 
\def \fs {\Mf(\Is,\thf)}
\def \ft {\Mf(\It,\thf)}

\def \Ys  {Y_S}   
\def \Yt  {Y_T}   
\def \yi  {y_i}   
\def \phii {\phi^c_i}
\def \Phin {\Phi^c_n}
\def \tPhin {\tilde{\Phi}^c_n}
\def \phiih {\phi^{\hat{c}}_i}
\def \tphii {\tilde{\phi}^c_i}

\def \pn {p_n} 
\def \pcn {p^c_n} 

\def \Rn {\mathcal{R}(n)} 

\def \C {\mathcal{C}} 
\def \Cstatic {\mathcal{C}_{static}} 
\def \Pstatic {\mathcal{P}_{static}} 

\section{Introduction}\label{sec.Intro}
Recent developments of technologies in computer vision, deep learning, and more broadly artificial intelligence, have led to the race of building advanced driver assistance systems (ADAS). From recognizing particular objects of interest toward understanding the corresponding driving environments, road scene segmentation is among the key components for a successful ADAS. With a sufficient amount of annotated training image data, existing computer vision algorithms already exhibit promising performances on the above task. However, when one applies pre-trained segmenters to a scene or city which is previously not seen, the resulting performance would be degraded due to dataset (domain) biases.


We conduct a pilot experiment to illustrate how severe a state-of-the-art semantic segmenter would be affected by the above dataset bias problem. We consider the segmenter of \cite{badrinarayanan2015segnet} which is trained on Cityscapes~\cite{cordts2016cityscape}, and apply for segmenting about 400 annotated road scene images of different cities across countries: Rome, Rio, Taipei, and Tokyo. A drop in mean of intersection over union (mIoU) of 25-30\% was observed (see later experiments for more details). Thus, how to suppress the dataset bias would be critical when there is a need to deploy road scene segmenters to different cities.

\begin{figure}
\begin{center}
\includegraphics[width=0.47\textwidth]{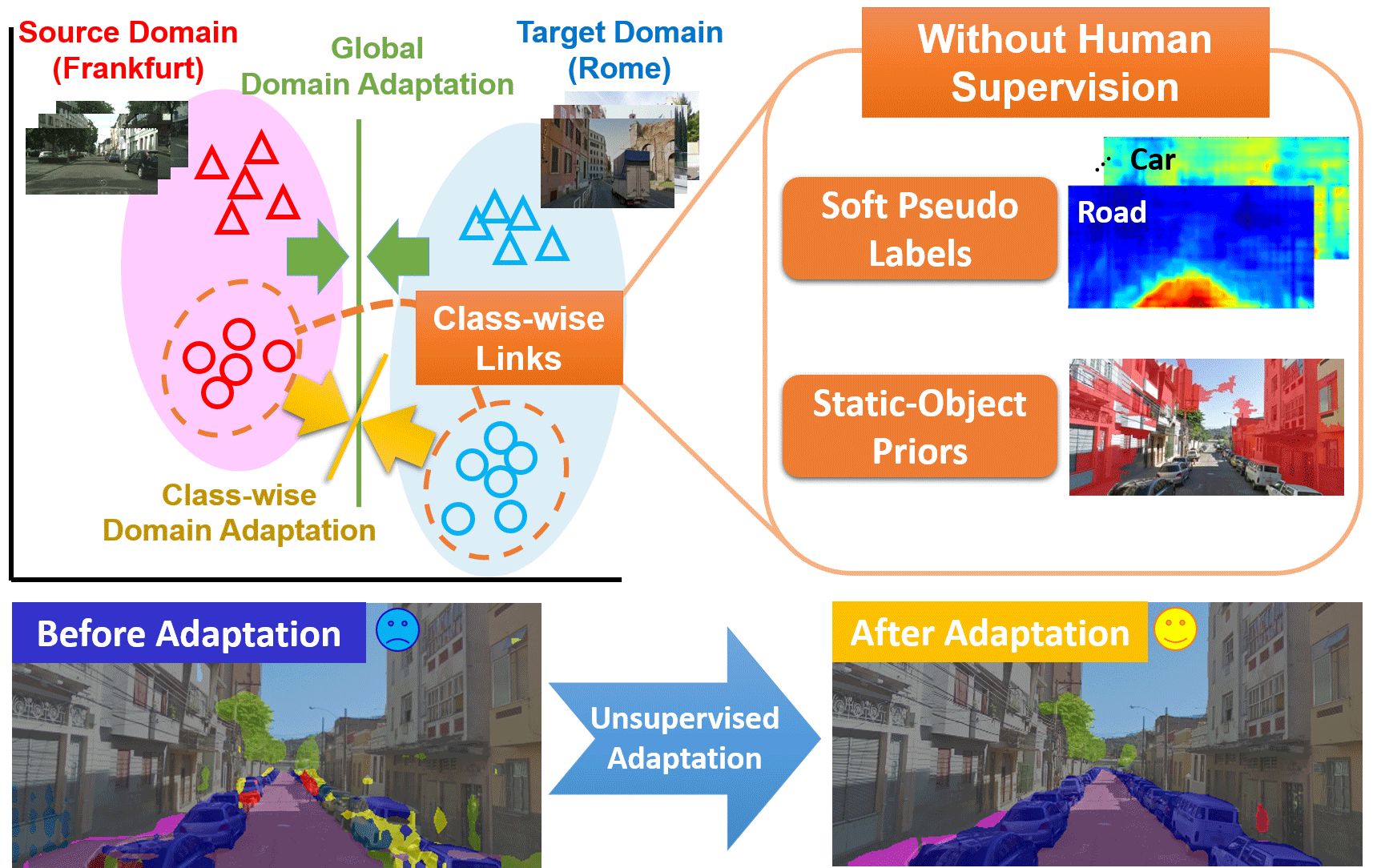}
\end{center}
\caption{Illustration of our \textit{unsupervised} domain adaptation method consisting of global and class-wise segmentation adaptations.
For class-wise adaptation, we leverage ``soft" pseudo labels and static object priors (obtained without human supervision) to further alleviate the domain discrimination in each class.}
\end{figure}

It is not surprising that, collecting a large number of annotated training image data for each city of interest would be time-consuming and expensive. For instance, pixel labeling of one Cityscapes image takes 90 minutes on average~\cite{cordts2016cityscape}. To alleviate this problem, a number of methods have been proposed to reduce human efforts in pixel-level semantic labeling. For example, researchers choose to utilize 3D information~\cite{xie2016semantic}, rendered images~\cite{richter2016playing,ros2016synthia}, or weakly supervised labels~\cite{saleh2016built,shimoda2016distinct,bearman2016s} for labeling. However, these existing techniques still require human annotation during data collection, and thus might not be easily scaled up to larger image datasets.

Inspired by the recent advances in domain adaptation~\cite{pan2010survey,torralba2011unbiased,khosla2012undoing}, we propose an unsupervised learning framework for performing cross-city semantic segmentation. Our proposed model is able to adapt a pre-trained segmentation model to a new city of interest, while only the collection of \textit{unlabeled} road scene images of that city is required. To avoid any human interaction or annotation during data collection, we utilize Google Street View with its time-machine\footnote[1]{https://maps.googleblog.com/2014/04/go-back-in-time-with-street-view.html} feature to harvest road scene images taken at the same (or nearby) locations but across different times. As detailed later in Sec.~\ref{sec.Tech}, this allows us to extract \textit{static-object priors} from the city of interest. By integrating such priors with the proposed global and class-specific domain adversarial learning framework, refining/adapting the pre-trained segmenter can be easily realized.



The main contributions of this paper can be summarized as follows:
\begin{itemize}
\item We propose an \textit{unsupervised} learning approach, which performs global and class-wise adaptation for deploying pre-trained road scene segmenters across cities.
\item We utilize Google Street View images with time-machine features to extract static-object priors from the collected image data, without the need of user annotation or interaction.
\item Along with the static-object priors, we advance adversarial learning for assigning pseudo labels to cross-city images, so that joint global and class-wise adaptation of segmenters can be achieved.

\end{itemize}

\section{Related Work}\label{sec.RW}

\subsection{CNN-based Semantic Segmentation}
Semantic segmentation is among the recent breakthrough in computer vision due to the development and prevalence of Convolutional Neural Networks (CNN), which has been successfully applied to predict dense pixel-wise semantic labels~\cite{farabet2013learning,long2015fully,noh2015learning,badrinarayanan2015segnet,chen2015semantic}. For example, Long et al.~\cite{long2015fully} utilize CNN for performing pixel-level classification, which is able to produce pixel-wise outputs of arbitrary sizes. In order to achieve high resolution prediction, \cite{noh2015learning,badrinarayanan2015segnet} further adapt deconvolution layers into CNN with promising performances. On the other hand, Chen et al.~\cite{chen2015semantic} choose to add a fully-connected CRF layer at their CNN output, which refines the pixel labels with context information properly preserved. We note that, since the goal of this paper is to adapt pre-trained segmenters across cities, we do not limit the use of particular CNN-based segmentation solvers in our proposed framework.

\subsection{Segmentation of Road Scene Images}
To apply CNN-based segmenters to road scene images, there are several attempts to train segmenters on large-scale image datasets~\cite{cordts2016cityscape,xie2016semantic,richter2016playing,ros2016synthia}. For example, Cordts et al.~\cite{cordts2016cityscape} release a natural road scene segmentation dataset, which consists of over $5000$ annotated images. Xie et al.~\cite{xie2016semantic} annotate 3D semantic labels in a scene, followed by  transferring the 3D labels into the associated 2D video frames. \cite{richter2016playing,ros2016synthia} collect semantic labels from Computer Graphic (CG) images at a large scale; however, building CG worlds for practical uses might still be computationally expensive. 
\begin{figure*}[t]
\begin{center}
\includegraphics[width=1.0\textwidth]
{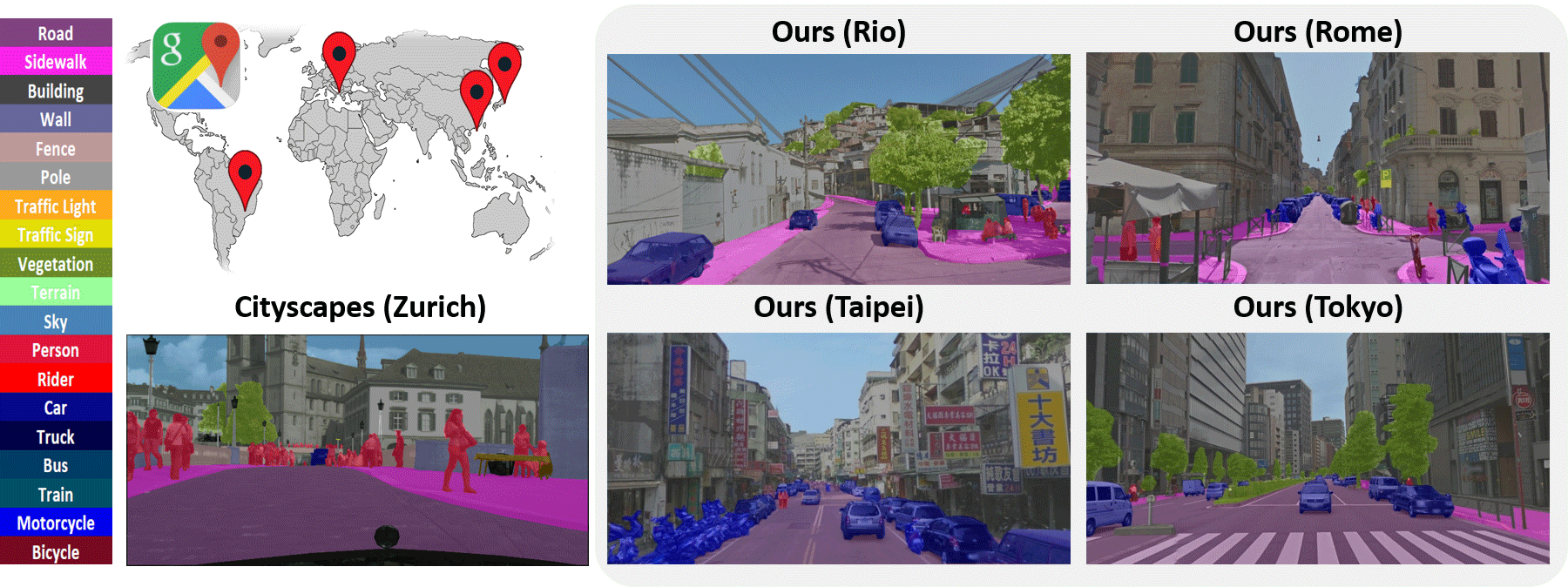}
\end{center}
\vspace{-0.2cm}
\caption{Example road scene images of different cities in our dataset. For evaluation purposes, we randomly select 100 images in each city to annotate pixel-level semantic labels. Color-coded labels are overlaid on each example image, where the mapping between colors and semantic classes are shown in the left panel.}
\label{fig.dataset}
\end{figure*}

On the other hand, \cite{bearman2016s} choose to relax the supervision during the data collection process, and simply require a number of point-labels per image. Moreover, \cite{papandreou2015weakly,pathak2015fully,pinheiro2015image} only require image-level labels during data collection and training. In addition to image-level labels, Pathak et al.~\cite{pathak2015constrained} incorporate constraints on object sizes, \cite{kolesnikov2016seed,shimoda2016distinct,saleh2016built} utilize weak object location knowledge, and \cite{kolesnikov2016seed} exploit object boundaries for constrained segmentation without using a large annotated dataset. Alternatively, \cite{lin2016scribblesup,xu2015learning} apply free-form squiggles to provide partial pixel labels for data collection. Finally, \cite{guillaumin2014imagenet} utilize image-level labels with co-segmentation techniques to infer semantic segmentation of foreground objects in the images of ImageNet.


\subsection{DNN-based Domain Adaptation}
Since the goal of our work is to adapt CNN-based segmenters across datasets (or cities to be more precise), we now review recent deep neural networks (DNN) based approaches for domain adaptation~\cite{pan2010survey}. Based on Maximum Mean Discrepancy (MMD), Long et al.~\cite{long2015learning} minimize the mean distance between data domains, and later they incorporate the concept of residual learning~\cite{long2016unsupervised} for further improvements. Zellinger et al.~\cite{zellinger2017central} consider Central Moment Discrepancy (CMD) instead of MMD, while Sener et al.~\cite{sener2016learning} enforce cyclic consistency on adaptation and structured consistency on transduction in their framework.

Recently, Generative Adversarial Network (GAN)~\cite{goodfellow2014generative} has raised great attention in the fields of computer vision and machine learning. While most existing architectures are applied for synthesizing images with particular styles~\cite{goodfellow2014generative, radford2015unsupervised,zhu2016generative}. Some further extend such frameworks for domain adaptation. In Coupled GAN~\cite{liu2016coupled}, domain adaptation is achieved by first generating corresponded instances across domains, followed by performing classification. 

In parallel with the appearance of GAN~\cite{goodfellow2014generative}, Ganin et al. propose Domain Adversarial Neural Networks (DANN)~\cite{ganin2015unsupervised,ganin2016domain}, which consider adversarial training for suppressing domain biases. For further extension, Variational Recurrent Adversarial Deep Domain Adaptation (VRADA)~\cite{purushotham2017variational} utilizes Variational Auto Encoder~(VAE) and RNN for time-series adaptation. Sharing a similar goal as ours, Hoffman et al.~\cite{hoffman2016fcns} extend such frameworks for semantic segmentation.

\section{Dataset}\label{sec.Dataset}
We now detail how we collect our road scene image dataset, and explain its unique properties.\\
\\
\noindent\textbf{Diverse locations and appearances.}
Using Google Street View, road scene images at a global scale can be accessed across a large number of cities in the world. To address the issue of geo-location discrimination of a road scene segmenter, we download the road scene images of four cities at diverse locations, Rome, Rio, Tokyo, and Taipei, which are expected to have significant appearance differences. To ensure that we cover sufficient variations in visual appearances from each city, we randomly sample the locations in each city for image collection.\\

\noindent\textbf{Temporal information.} With the time-machine features of Google Street View, image pairs of the same location yet across different times can be further obtained. As detailed later in the Sec.~\ref{CA}, this property particularly allows us to observe prior information from static objects, so that improved adaptation without any annotation can be achieved. In our work, we have collected 1600 image pairs (3200 images in total) at 1600 different locations per city with high image quality ($647\times1280$ pixels).
    
For evaluation purposes, we select 100 image pairs from each city as the testing set, with pixel-level ground truth labels annotated by 15 image processing experts. We define 13 major classes for annotation: road, sidewalk, building, traffic light, traffic sign, vegetation, sky, person, rider, car, bus, motorcycle, and bicycle, as defined in Cityscapes~\cite{cordts2016cityscape}. Fig.~\ref{fig.dataset} shows example images of our dataset. The dataset will be publicly available later for academic uses. To see more details and examples of our dataset, please refer to Appendix~\ref{Appendix_2} or visit our website: 
\urlstyle{rm}\url{https://yihsinchen.github.io/segmentation_adaptation/}.\\

We now summarize the uniqueness of our dataset below:
\begin{itemize}
\item Unlike existing datasets which typically collect images in nearby locations (e.g., road scenes of the same city), our dataset includes over 400 road scene images from four different cities around the world, with high-quality pixel-level annotations (for evaluation only).
\item Our dataset include image pairs at the same location but across different times, which provide additional temporal information for further processing and learning purposes.
\end{itemize}

\begin{figure*}[t]
\begin{center}
\includegraphics[width=1.0\textwidth]
{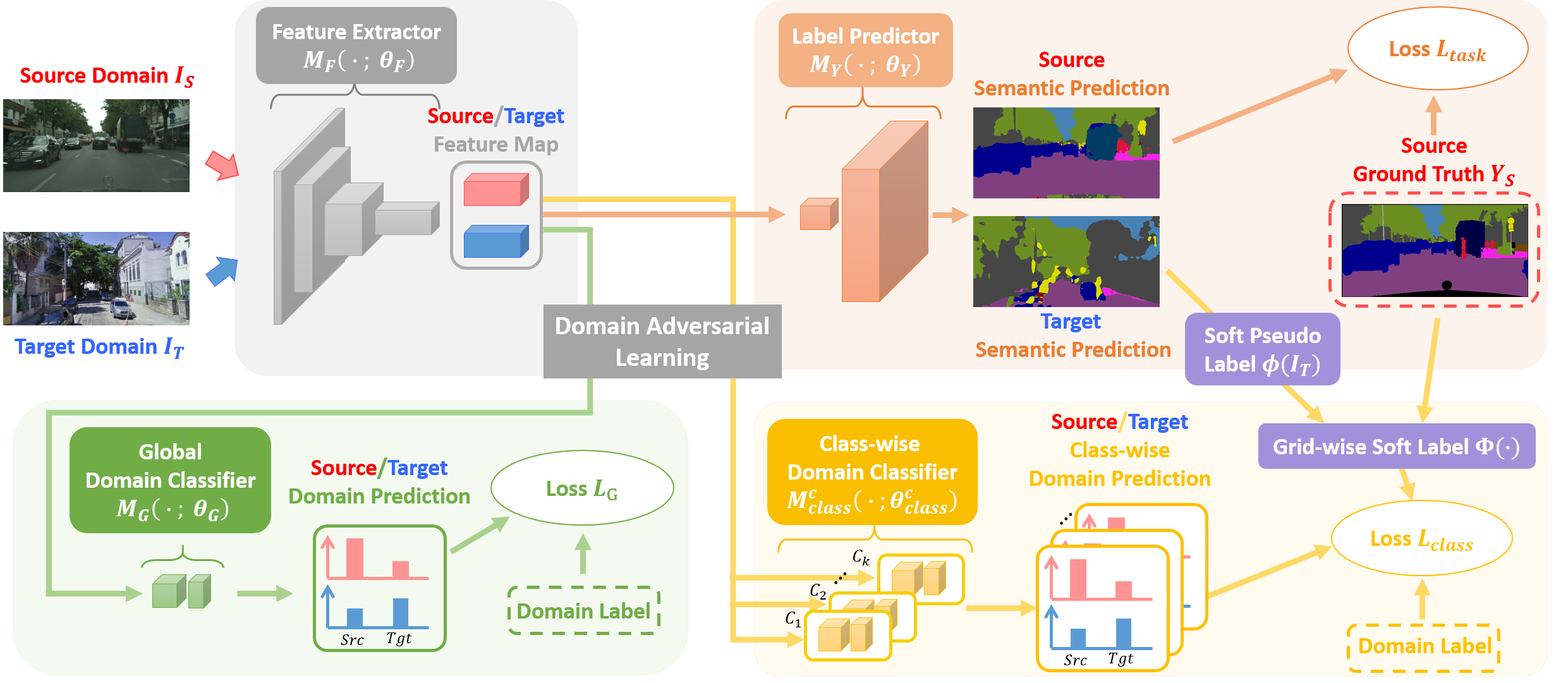}
\end{center}
\vspace{-3mm}
\caption{\small The overview of our proposed DNN framework. The feature extractor $\Mf$ transforms cross-domain images into a proper feature space, which is derived by performing global $\Mga$ and class-wise $\Mca$ domain alignment  via adversarial learning. The label predictor $\My$ regularizes the learned model by only observing the ground-truth annotation of source-domain images.}
\label{fig.framework}
\end{figure*}

\section{Our Method}\label{sec.Tech}
In this section, we present the details of our proposed \textit{unsupervised domain  adaptation} framework, which is able to adapt pre-trained segmenters across different cities without using any user annotated data. In other words, while both images $\Is$ and labels $\Ys$ are available from the source domain $\S$, only images $\It$ for the target domain $\T$ can be observed.\\

\noindent\textbf{Domain shift.} When adapting image segmenters across cities, two different types of domain shifts (or dataset biases) can be expected: \textit{global} and \textit{class-wise domain shift}. The former comes from the overall differences in appearances between the cities, while the latter is due to distinct compositions of road scene components in each city.

To minimize the global domain shift, we follow~\cite{hoffman2016fcns} and apply the technique of adversarial learning, which introduces a domain discriminator with a loss $\Lga$. This is to distinguish the difference between source and target-domain \textit{images}, with the goal to produce a common feature space for images across domains. To perform class-wise alignment, we extend the above idea and utilize multiple \textit{class-wise} domain discriminators (one for each class) with the corresponding adversarial loss $\Lca$. Unlike the discriminator for global alignment, these class-wise discriminators are trained to suppress the difference between cross-domain images but of the same class. Since we do not have any annotation for the city of interest (i.e., target-domain images), later we will explain how our method performs unsupervised learning to jointly solve the above adaptation tasks.

With the above loss terms defined, the overall loss of our approach can be written as: 
\begin{equation}\label{loss_stat}
	\Ltotal = \Ltask + \lmga\Lga + \lmca\Lca ~,
\end{equation}
where $\lmga$ and $\lmca$ are weights for the global and class-wise domain adversarial loss, respectively. Note that $\Ltask$ denotes the prediction loss of source-domain images, which can be viewed as a regularization term when adapting the learned model across domains.\\

\noindent\textbf{Our proposed framework.} Fig.~\ref{fig.framework} illustrates our framework. Let $\C$ be the set of classes, and an input image denoted as $x$. Our proposed architecture can be decoupled into four major components: feature extractor $\fx$ that transforms the input image to a high-level, semantic feature space (the gray part), label predictor $\My(\fx, \thy)$ that maps feature space to task label space (the orange part), and domain discriminator for global $\Mga(\fx, \thga)$ (the green part) and class-wise $\Mca(\fx, \thca), c \in \C$  alignments (the yellow part). The feature extractor and task label predictor are initialized from a pre-trained segmenter, while the domain discriminators are randomly initialized. While we utilize the front-end dilated-FCN~\cite{yu2016multi} as the pre-trained segmenter in our work, it is worth noting that our framework can be generally applied to other semantic segmenters. 

In Sec.~\ref{GA} and Sec.~\ref{CA}, we will detail our unsupervised learning for global alignment and class-wise alignment, respectively. In particular, how we extract and integrate static-object priors for the target domain images without any human annotation will be introduced in Sec.~\ref{prior}.

\subsection{Global Domain Alignment}\label{GA}

Previously, domain adversarial learning frameworks have been applied for solving cross-domain image classification tasks~\cite{ganin2015unsupervised}. However, for cross-domain image segmentation, each image consists of multiple pixels, which can be viewed as multiple instances per observation. Thus, how to extend the idea of domain adversarial learning for adapting segmenters across image domains would be our focus.

Inspire by~\cite{hoffman2016fcns}, we take each \textit{grid} in the $fc7$ feature map of the FCN-based segmenter as an instance. Let the feature maps of source and target domain images as $\Mf(\Is, \thf)$ and $\Mf(\It, \thf)$, each map consists of $N$ grids. Let $\pn(x)=\sigma(\Mga(\fx_n, \thga))$ be the probability that the grid $n$ of image $x$ belongs to the source domain, where $\sigma$ is the sigmoid function. We note that, for cross-domain classification, Ganin et al.~\cite{ganin2015unsupervised} use the same loss function plus a gradient reversal layer to update the feature extractor and domain discriminator simultaneously. If directly applying their loss function for cross-domain segmentation, we would observe:

\begin{align}\label{reverse_grad}
\max_{\thf}\min_{\thga}~~\Lga = 
& -\sum_{\Is\in\S}\sum_{n\in N} log(\pn(\Is)) \nonumber\\
& - \sum_{\It\in\T}\sum_{n\in N} log(1-\pn(\It))~.
\end{align}

Unfortunately, this loss function will result in gradient vanishing as the discriminator converges to its local minimum. To alleviate the above issue, we follow~\cite{goodfellow2014generative} and decompose the above problem into two subtasks. More specifically, we have a domain discriminator $\thga$ trained with $\Lga^{D}$ for classifying these two distributions into two groups, and a feature extractor $\thf$ updated by its inverse loss $\Lga^{Dinv}$ which minimizes the associated distribution differences. In summary, our objective is to minimize $\Lga = \Lga^{D} + \Lga^{Dinv}$ by iteratively update $\thga$ and $\thf$:
\begin{align}\label{train_GA}
\min_{\thga}~~~\Lga^{D}  
\textrm{ ,   }
\min_{\thf}~~~\Lga^{Dinv}~,
\end{align}
where $\Lga^{D}$ and $\Lga^{Dinv}$ are defined as:
\begin{align}
\Lga^{D} = & -\sum_{\Is\in \S}\sum_{n\in N} log(\pn(\Is)) \nonumber\\
& - \sum_{\It\in\T}\sum_{n\in N} log(1-\pn(\It))~,
\end{align}
\begin{align}
\Lga^{Dinv} = 
& -\sum_{\Is\in\S}\sum_{n\in N}\ log(1-\pn(\Is)) 
\nonumber\\
& - \sum_{\It\in\T}\sum_{n\in N} log(\pn(\It))~.
\end{align}

\subsection{Class-wise Domain Alignment}\label{CA}

In addition to suppressing the global misalignment between image domains, we propose to advance the same adversarial learning architecture to perform class-wise domain adaptation. 

While the idea of regularizing class-wise information during segmenter adaptation has been seen in~\cite{hoffman2016fcns}, its class-wise alignment is performed based on the composition of the class components in cross-city road scene images. To be more precise, it assumes that the composition/proportion of object classes across cities would be similar. Thus, such a regularization essentially performs global instead of class-specific adaptation.

Recall that, when adapting our segmenters across cities, we only observe road scene images of the target city of interest without any label annotation. Under such unsupervised settings, we extend the idea in~\cite{long2013transfer} and assign pseudo labels to pixels/grids in the images of the target domain. That is, after the global adaptation in Fig.~\ref{fig.framework}, the predicted probability distribution maps $\phi(\It) = \textrm{softmax}(\My(\ft, \thy))$ of target domain images can be produced. Thus, $\phi(\It)$ can be viewed as the ``soft" pseudo label map for the target domain images. As a result, class-wise association across data domains can be initially estimated by relating the ground truth label in the source domain and the soft pseudo label in the target domain.\\

\noindent\textbf{From pixel to grid-level pseudo label assignment.} 
In Sec.~\ref{GA}, to train the domain discriminator, we define each grid $n$ in the feature space as one instance, which corresponds to multiple pixels in the image space. If the (pseudo) labels of these grids can be produced, adapting class-wise information using the same adversarial learning framework can be achieved.

To propagate and to determine the pseudo labels from pixels to each grid for the above adaptation purposes, we simply calculate the proportion of each class in each grid as the \textit{soft} (pseudo) label. That is, let $i$ be the pixel index in image space, $n$ be the grid index in feature space, and $\Rn$ be the set of pixels that correspond to grid $n$. If $\yi(\Is)$ denote the ground truth label of pixel $i$ for source domain images, we then calculate source-domain grid-wise soft-label $\Phin(\Is)$ as the probability of grid $n$ belonging to class~$c$:
\begin{equation}\label{sourceweighting}
\Phin(\Is) = \sum_{i \in\Rn}\frac{\yi(\Is)==c}{\mid\Rn\mid}.
\end{equation}

On the other hand, due to the lack of annotated target-domain data, it is not as straightforward to assign grid-level soft pseudo labels to images in that domain. To solve this problem, we utilize $\phi(I_T)$ derived above. Let $\phii(I_T)$ be the pixel-wise soft pseudo label of pixel $i$ corresponding to class $c$ for target-domain images, we have target grid-wise soft pseudo label $\Phin(\It)$ of grid~$n$:
\begin{equation}\label{targetweighting}
\Phin(\It) = \sum_{i \in \mathcal{R}(n)}\frac{\phii(I_T)}{\mid\mathcal{R}(n)\mid}~.
\end{equation}

Intuitively, grid-wise soft (pseudo) labels $\Phin(\Is)$ and $\Phin(\It)$ are estimations of the probabilities that each grid $n$ in source and target domain images belongs to object class $c$. To balance the appearance frequency of different classes, we normalize the estimated outputs in~\eqref{sourceweighting} and~\eqref{targetweighting} as follows:

\begin{align}\label{normalized_soft-label}
	&\tPhin(\Is) = \frac{\Phin(\Is)}{\sum\limits_{n\in N} \Phin(\Is)}~
	\nonumber\\
	&\tPhin(\It) = \frac{\Phin(\It)}{\sum\limits_{n\in N} \Phin(\It)}~.
\end{align}
\begin{figure}[t]
\begin{center}
\includegraphics[width=0.4\textwidth]{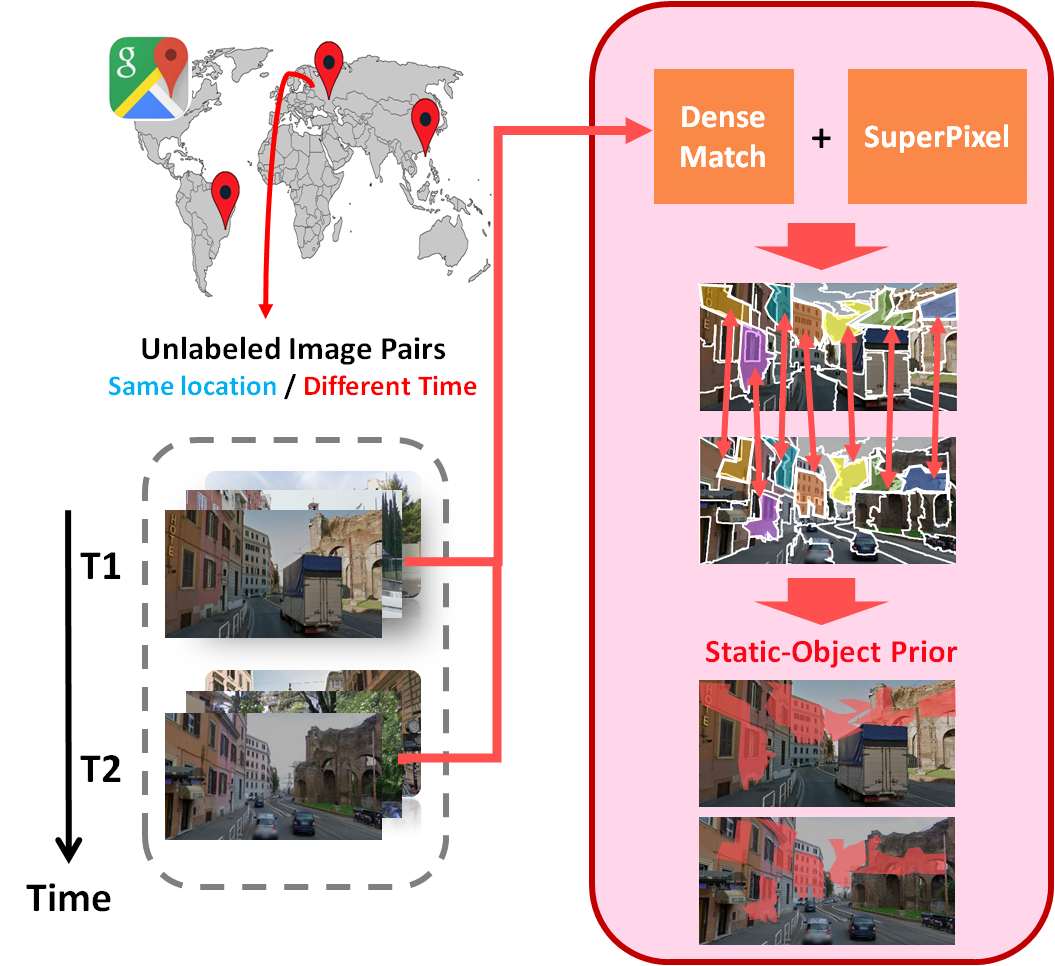}
\end{center}
\vspace{-2mm}
\caption{\small Illustration of static-object prior extraction. Given a pair of images at the same location but at different times, image regions belonging to static objects (e.g., the red blobs) can be identified by performing dense matching and superpixel segmentation.}
\label{fig:static_pipeline}
\end{figure}

\noindent\textbf{Class-wise adversarial learning.}
With the soft labels assigned to the source-domain images and the soft pseudo labels predicted for the target-domain ones, we now explain our adversarial learning for class-wise domain adaptation.

As depicted in Fig.~\ref{fig.framework}, we deploy multiple \textit{class-wise} domain discriminators $\thca, c \in \C$ in our proposed architecture, and each discriminator is specially trained for differentiating \textit{objects} of the corresponding class $c$ across domains. Similar to $\pn(x)$, given that each object class $c$ has a corresponded domain discriminator $\Mca$, we define $\pcn(x)=\sigma(\Mca(\fx_n, \thca))$ as the probability predicted by $\Mca$ that the grid $n$ of image $x$ is from the source domain. Combining the definition in \eqref{normalized_soft-label}, we define a pair of class-wise adversarial loss $\Lca^{D}$ and $\Lca^{Dinv}$ to guide the optimization for class-wise alignment:


\begin{align}\label{loss_CA}
	\Lca^{D} = 
    &-\sum_{\Is\in\S}\sum_{c\in\C}\sum_{n\in N}\tPhin(\Is)log(\pcn(\Is)) 
    \nonumber\\
    &-\sum_{\It\in\T}\sum_{c\in\C}\sum_{n\in N}\tPhin(\It)log(1-\pcn(\It))~,
\end{align}
\begin{align}\label{loss_CA_inv}
	\Lca^{Dinv} = &-\sum_{\Is\in\S}\sum_{c\in\C}\sum_{n\in N}\tPhin(\Is)log(1-\pcn(\Is)) 	
    \nonumber\\
    &-\sum_{\It\in\S}\sum_{c\in\C}\sum_{n\in N}\tPhin(\It)log(\pcn(\It))~.
\end{align}

Finally, similar to~\eqref{train_GA}, the class-wise alignment process is to iteratively solve the following optimization problem:
\begin{align}\label{train_CA}
\min_{\bigcup\limits_{c \in \C} \thca}~\Lca^{D} \textrm{ , }
\min_{\thf}~~~\Lca^{Dinv}~,
\end{align}
which minimizes the overall loss $\Lca = \Lca^{D} + \Lca^{Dinv}$~.

\subsection{Harvesting Static-Object Prior}\label{prior}
While jointly performing global and class-wise alignment between source and target-domain images would produce promising adaptation performance, the pseudo labels are initialized by pre-trained segmenter. Under the unsupervised domain adaptation setting, since no annotation of target-domain data can be obtained, fine-tuning the segmenter by such information is not possible.

However, with the use of time-machine features from Google Street View images, we are able to leverage the temporal information for extracting the static-object priors from images in the target domain. As illustrated in Fig.~\ref{fig:static_pipeline}, given an image pair of the same location but across different times, we first apply DeepMatching~\cite{weinzaepfel2013deepflow} to relate pixels within each image pair. For the regions with matched pixels across images, it implies such regions are related to static objects (e.g., building, road, etc.). Then, we additionally perform superpixel segmentation on the image pair using Entropy Rate Superpixel~\cite{liu2011entropy}, which would group the nearby pixels into regions while the boundaries of the objects can be properly preserved. With the above derivation, we view the matched superpixels containing more than $k$ matched pixels (we fix $k=3$ in this work) as the static-object prior $\Pstatic(\It)$. Please refer to Appendix~\ref{Appendix_1} for typical examples of mining static-object prior.

Let $\Cstatic$ be the set of static-object classes. For the pixels that belong to $\Pstatic(\It)$, we then refine their soft pseudo labels by suppressing its probabilities of being non-static objects:
\begin{align}\label{suppress_normalize}
\forall~i &\in \Pstatic(\It) \nonumber\\
\tphii(\It) &=
\begin{cases}
~\phii(\It)~/{\sum\limits_{\hat{c} \in \Cstatic}\phiih(\It)} & \quad \text{if } c \in \Cstatic \\
~~~~~~~~~~~~0 & \quad \text{else} 
\end{cases}
\end{align}




\begin{table}[t]
\caption{\small Accuracy of applying dilated-FCNs pre-trained on Cityscapes (Frankfurt) to different cities (i.e., no adaptation).}
\begin{center}
\begin{tabular}{lll}
\hline
\textbf{City} & \textbf{Dataset} & \textbf{mIOU (\%)} \\
\hline
Frankfurt & Cityscapes & 64.6\% \\\hline
Rome	 & Ours & 38.2\% \\
Tokyo	 & Ours & 39.2\% \\
Rio & Ours & 38.5\% \\
Taipei & Ours & 35.1\% \\\hline
\end{tabular}
\end{center}
\label{tab.dis}
\vspace{-5mm}
\end{table}

\begin{table*}[t!]\vspace{-5mm}
\caption{\small Segmentation performance comparisons (in mIOU), in which SW, BLDG, TL, TS, VEG, Motor stand for Sidewalk, Building, Traffic Light, Traffic Sign, Vegetation, and Motorbike, respectively. Note that GA/CA denote the components of global/class-wise adaptation in our architecture, while our method (Full Method) integrates both components with static-object priors for unsupervised domain adaptation. The performance upper bound achieved by the fully supervised baseline is noted as UB.}
	\begin{adjustbox}{width=1.0\textwidth}
    \small
	\begin{tabular}{|c|c|c|c|c|c|c|c|c|c|c|c|c|c|c|c|}
		\hline    
        \multicolumn{1}{|c|}{\multirow{2}{*}{City}} & \multicolumn{1}{c|}{\multirow{2}{*}{Method}} & \multicolumn{14}{c|}{\textbf{Cityscapes $\to$ Our Dataset}}\\ \cline{3-16}
    	& & Road & SW & BLDG & TL & TS & VEG & Sky & Person & Rider & Car & Bus & Motor. & Bicycle & mIOU\\\hline   	
        \multicolumn{1}{|c|}{\multirow{5}{*}{Rome}}&Pre-trained & 77.7 & 21.9 & 83.5 & 0.1 & 10.7 & 78.9 & \textbf{88.1} & 21.6 & 10.0 & 67.2 & 30.4 & 6.1 & 0.6 & 38.2 \\ \cline{2-16}
    	
    	&GA & 79.2 & 25.7 & 84.0 & \textbf{0.1} & 11.8 & 81.0 & 83.3 & 29.3 & 8.9 & 71.8 & 35.9 & 23.7 & 0.9 & 41.2\\ 
    	&GA+CA & 78.2 & 26.0 & \textbf{84.9} & 0.0 & 21.5 & \textbf{81.7} & 83.0 & \textbf{31.0} & 11.2 & \textbf{72.0} & 33.0 & 24.1 & \textbf{1.2} & 42.1\\ 
    	&Full Method & \textbf{79.5} & \textbf{29.3} & 84.5 & 0.0 & \textbf{22.2} & 80.6 & 82.8 & 29.5 & \textbf{13.0} & 71.7 & \textbf{37.5} & \textbf{25.9} & 1.0 & \textbf{42.9}\\ \cline{2-16}
        &\textit{UB} & 84.9 & 33.0 & 87.3 & 0.0 & 10.9 & 84.6 & 91.6 & 30.5 & 19.1 & 77.7 & 10.6 & 38.3 & 0.5 & 43.8\\ \hline   \hline	
        \multicolumn{1}{|c|}{\multirow{5}{*}{Rio}}&Pre-trained & 69.0 & 31.8 & 77.0 & \textbf{4.7} & 3.7 & 71.8 & \textbf{80.8} & 38.2 & 8.0 & 61.2 & 38.9 & 11.5 & 3.4 & 38.5\\ \cline{2-16}
    	&GA & 72.8 & 42.2 & 79.0 & 4.4 & 6.1 & 76.2 & 75.3 & 38.9 & 7.1 & 66.5 & 41.2 & 16.9 & 5.5 & 40.9\\ 
    	&GA+CA & 73.2 & 42.9 & 78.4 & 3.3 & \textbf{7.9} & 76.2 & 72.4 & 39.1 & 9.1 & \textbf{68.3} & \textbf{43.8} & 16.8 & 6.5 & 41.4\\ 
    	&Full Method & \textbf{74.2} & \textbf{43.9} & \textbf{79.0} & 2.4 & 7.5 & \textbf{77.8} & 69.5 & \textbf{39.3} & \textbf{10.3} & 67.9 & 41.2 & \textbf{27.9} & \textbf{10.9} & \textbf{42.5}\\ \cline{2-16}
        &\textit{UB} & 80.2 & 53.8 & 84.5 & 0.0 & 16.4 & 81.4 & 85.4 & 42.3 & 17.4 & 74.0 & 49.4 & 37.3 & 16.7 & 49.1\\ \hline \hline  	
        \multicolumn{1}{|c|}{\multirow{5}{*}{Tokyo}}&Pre-trained & 81.2 & 26.7 & 71.7 & 8.7 & 5.6 & 73.2 & \textbf{75.7} & 39.3 & 14.9 & 57.6 & 19.0 & 1.6 & 33.8 & 39.2\\ \cline{2-16}
    	&GA & 83.5 & \textbf{36.2} & 72.3 & 10.8 & 7.1 & 77.0 & 66.2 & 44.0 & 18.6 & 61.5 & \textbf{21.9} & 4.9 & 37.5 & 41.7\\ 
    	&GA+CA & \textbf{83.6} & 36.1 & 71.9 & 11.3 & \textbf{13.0} & \textbf{77.6} & 64.4 & 41.2 & 19.3 & 63.7 & 20.2 & \textbf{13.9} & 38.8 & 42.6\\ 
    	&Full Method & 83.4 & 35.4 & \textbf{72.8} & \textbf{12.3} & 12.7 & 77.4 & 64.3 & \textbf{42.7} & \textbf{21.5} & \textbf{64.1} & 20.8 & 8.9 & \textbf{40.3} & \textbf{42.8}\\ \cline{2-16}
        &\textit{UB} & 85.2 & 38.7 & 79.8 & 13.9 & 19.7 & 81.7 & 86.9 & 45.3 & 35.9 & 66.9 & 29.0 & 2.0 & 42.4 & 48.3\\\hline\hline 
        \multicolumn{1}{|c|}{\multirow{5}{*}{Taipei}}&Pre-trained & 77.2 & 20.9 & 76.0 & 5.9 & 4.3 & 60.3 & 81.4 & 10.9 & \textbf{11.0} & 54.9 & 32.6 & 15.3 & 5.2 & 35.1\\ \cline{2-16}
    	&GA & 79.0 & 27.7 & 76.6 & 13.1 & 5.0 & 67.7 & 74.8 & \textbf{17.5} & 6.1 & 60.4 & 28.9 & 25.5 & 7.1 & 37.6\\ 
    	&GA+CA & \textbf{79.2} & \textbf{29.0} & \textbf{80.3} & \textbf{14.1} & \textbf{8.2} & \textbf{68.8} & 81.1 & 16.3 & 10.5 & \textbf{64.7} & 33.8 & 16.2 & 6.5 & 38.8\\ 
    	&Full Method & 78.6 & 28.6 & 80.0 & 13.1 & 7.6 & 68.2 & \textbf{82.1} & 16.8 & 9.4 & 60.4 & \textbf{34.0} & \textbf{26.5} & \textbf{9.9} & \textbf{39.6}\\ \cline{2-16}
        &\textit{UB} & 84.0 & 36.6 & 87.7 & 9.9 & 13.7 & 76.2 & 91.9 & 23.4 & 24.1 & 65.1 & 39.4 & 47.8 & 3.2 & 46.4\\\hline   	
	\end{tabular}
    \end{adjustbox}
 	\label{tab.exp}
\end{table*}

\section{Experiments}\label{sec.Exp}
We first conduct experiments to demonstrate the issue of cross-city discrimination even using a state-of-the-art semantic segmenter. Then, we will verify the effectiveness of our proposed \textit{unsupervised learning} method on the \textbf{Cityscapes to Our Dataset} domain adaptation task. By comparing it with a fully-supervised baseline (i.e., fine-tuning by fully annotated training data), we show that our unsupervised method would achieve comparable performances as the fully-supervised methods in most cases. Finally, we perform an extra experiment, \textbf{SYNTHIA to Cityscapes}, to prove that our method could be generally applied to different datasets.
\subsection{Implementation Details}
In this work, all the implementations are produced utilizing the open source TensorFlow \cite{abadi2016tensorflow} framework, and the codes will be released upon acceptance. In the following experiments, we use mini-batch size $16$ and the Adam optimizer \cite{kingma2015adam} with learning rate of $5\times10^{-6}$, $beta1=0.9$, and $beta2=0.999$ to optimize the network. Moreover, we set the hyper-parameters in \eqref{loss_stat}: $\lmga$ and $\lmca$, to be numbers gradually changing from 0 to 0.1 and 0 to 0.5, respectively. In addition, for the experiments using static-object priors, we use \{road, sidewalk, building, wall, fence, pole, traffic light, traffic sign, vegetation, terrain, sky\} as the set of static-object classes $\Cstatic$ defined in Sec.~\ref{prior}.

\subsection{Cross-City Discrimination}\label{dis}
We apply the segmenter pre-trained on \textbf{Cityscapes} to images of different cities in \textbf{Our Dataset}. As shown in Table~\ref{tab.dis}, there is a severe performance drop in the four cities compared to its original performance on Cityscapes. Interestingly, we observe a trend that the farther the geo-distance between the target city and the pre-trained city (Frankfurt), the severer the performance degradation. This implies that different visual appearances across cities due to cultural differences would dramatically impact the accuracy of the segmenter. For example, in Taipei, as shown in Fig.~\ref{fig.dataset}, there are many signboards and shop signs attached to the buildings, and many scooters on the road, which are uncommon in Frankfurt. It also justifies the necessity of an effective domain adaptation method for the road scene segmenter to alleviate the discrimination.

\begin{figure*}[t!]\vspace{-1mm}
\begin{center}
\includegraphics[width=1.0\textwidth]
{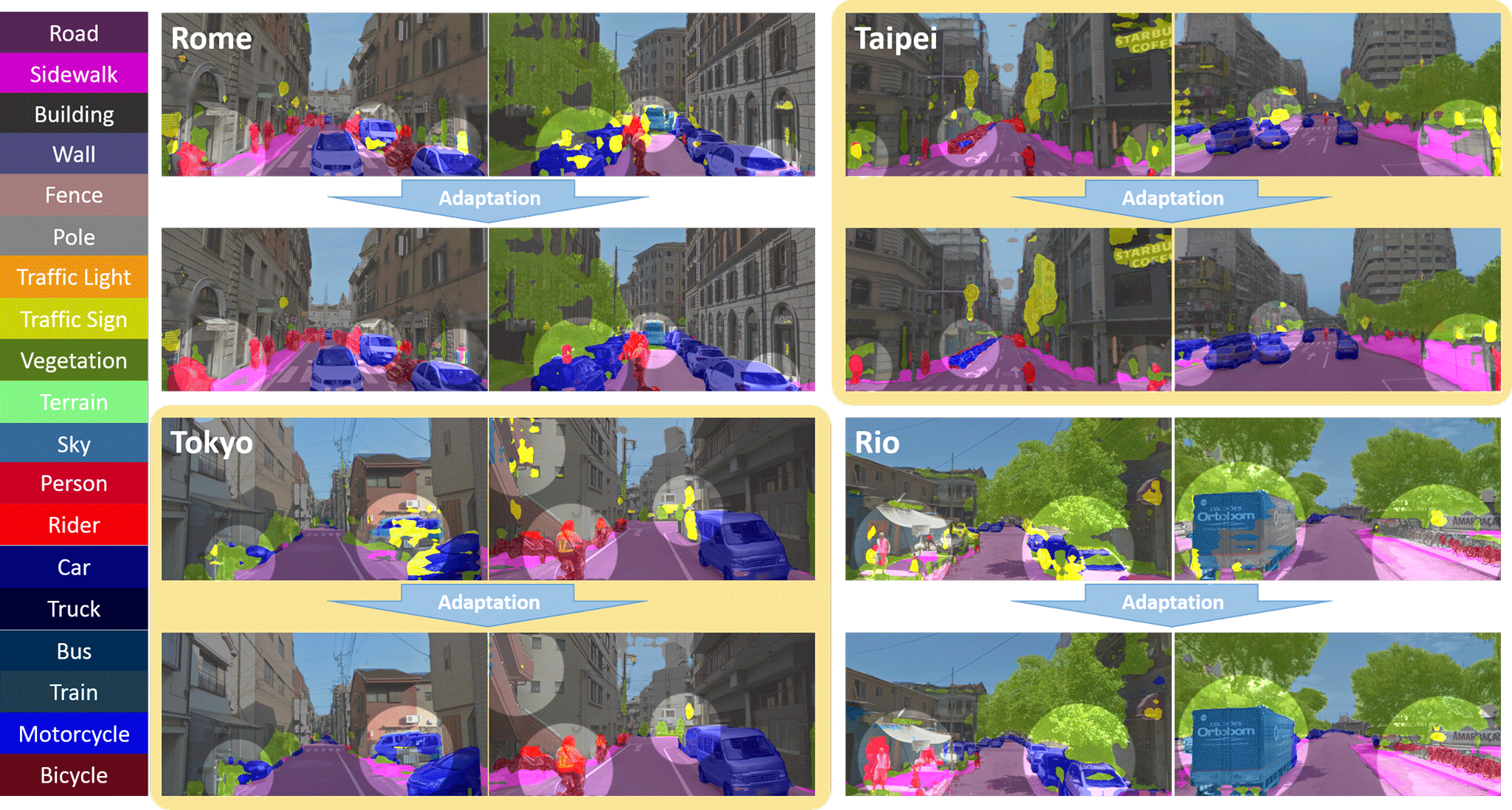}
\end{center}
\vspace{-2mm}
\caption{\small Examples of cross-city adaptation. The first/third and second/fourth rows show the results before and after adaptation, respectively. The regions with improved segmentation adaptation are highlighted for better visualization.}
\label{fig.typical_example}
\vspace{+1mm}
\end{figure*}

\begin{table*}[t!]
\caption{\small Experimental results for the SYNTHIA-to-Cityscapes segmentation adaptation task.}
\begin{adjustbox}{width=1.0\textwidth}
\small
\begin{tabular}{|c|c|c|c|c|c|c|c|c|c|c|c|c|c|c|}
\hline
\multirow{2}{*}{Method} & \multicolumn{14}{c|}{\textbf{SYNTHIA $\to$ Cityscapes}}                                       \\ \cline{2-15} 
                        & Road & SW & BLDG & TL & TS & VEG & Sky & Person & Rider & Car & Bus & Motor. & Bicycle & mIOU \\ \hline
Pre-trained & 24.3 & 19.5 & 48.3 & \textbf{1.5} & 5.4 & 77.4 & 76.1 & \textbf{42.8} & \textbf{9.7} & 62.5 & 9.8 & 0.5 & \textbf{20.9} & 30.7     \\ 
\hline
GA & 56.5 & 24.0 & \textbf{78.9} & 1.1 & \textbf{5.9} & 77.8 & 77.3 & 35.8 & 5.4 & 61.7 & 5.2 & 0.9 & 8.4 & 33.8     \\ 
GA+CA & \textbf{62.7} & \textbf{25.6} & 78.3 & 1.2 & 5.4 & \textbf{81.3} & \textbf{81.0} & 37.4 & 6.4 & \textbf{63.5} & \textbf{16.1} & \textbf{1.2} & 4.6 & \textbf{35.7}     \\ \hline
\end{tabular}
\end{adjustbox}
\label{tab.exp_syn_to_real}
\end{table*}

\subsection{Cross-City Adaptation}

\noindent\textbf{Baseline.} We use a \textit{fully-supervised} method to establish a strong baseline as the upper bound of adaptation improvement. We divide our 100 images with fine annotations to 10 subsets for each city. Each time we select one subset as the testing set, and the other 90 images as the training set and fine-tune the segmenter for 2000 steps. We repeat the procedure for $10$ times and average the testing results as the baseline performance.

\noindent\textbf{Our method.} Now we apply our domain adversarial learning method to adapt the pre-segmenter in an unsupervised fashion. Meanwhile, we do the ablation study to demonstrate the contribution from each component: global alignment, class-wise alignment, and static-object prior. We summarize the experimental results in Table~\ref{tab.exp}, where "Pre-trained" denotes the pre-trained model, "UB" denotes the \textit{fully-supervised} upper bound, "GA" denotes the global alignment part of our method, "GA+CA" denotes the combination of global alignment and class-wise alignment, and finally, "Full Method" denotes our overall method that utilizes the static-object priors. On average over four cities, our global alignment method contributes 2.6\% mIoU gain, our class-wise alignment method also contributes 0.9\% mIoU gain, and finally, the static-object priors contributes another 0.6\% mIOU improvement. Furthermore, the t-SNE visualization results in Appendix~\ref{Appendix_1} also show that the domain shift keeps decreasing from "Pre-trained" to "GA" to "GA+CA". These results demonstrate the effectiveness of each component of our method. In Fig.~\ref{fig.typical_example}, we show some typical examples.

\subsection{Synthetic to Real Adaptation}\label{syn2real}

We additionally apply our method to another adaptation task with a different type of domain shift: \textbf{SYNTHIA to Cityscapes}. In this experiment, we take SYNTHIA-RAND-CITYSCAPES~\cite{ros2016synthia} as the source domain, which contains 9400 synthetic road scene images with Cityscapes-compatible annotations. For the unlabeled target domain, we use the training set of Cityscapes. During evaluation, we test our adapted segmenter on the validation set of Cityscapes. We note that, since there are no paired images with temporal information in Cityscapes (as those in our dataset), we cannot extract static-object priors in this experiment. Nevertheless, from the results shown in Table~\ref{tab.exp_syn_to_real}, performing global and class-wise alignment using our proposed method still achieves 3.1\% and 1.9\% mIOU gain, respectively. These results again demonstrate the robustness of our proposed method. For typical examples of this adaptation task, please refer to Appendix~\ref{Appendix_3}.

\section{Conclusion}\label{sec.Con}

In this paper, we present an \textit{unsupervised} domain adaptation method for semantic segmentation, which alleviates cross-domain discrimination on road scene images across different cities. We propose a unified framework utilizing domain adversarial learning, which performs joint global and class-wise alignment by leveraging soft labels from source and target-domain data. In addition, our method uniquely identifies and introduce static-object priors to our method, which are retrieved from images via natural synchronization of static objects over time. Finally, we provide a new dataset containing road scene images of four cities across countries, good-quality annotations and paired images with temporal information are also included. We demonstrate the effectiveness of each component of our method on tasks with different levels of domain shift.

{\small
\bibliographystyle{ieee}

\begin{thebibliography}{10}\itemsep=-1pt

\bibitem{abadi2016tensorflow}
M.~Abadi, A.~Agarwal, P.~Barham, E.~Brevdo, Z.~Chen, C.~Citro, G.~S. Corrado,
  A.~Davis, J.~Dean, M.~Devin, et~al.
\newblock Tensorflow: Large-scale machine learning on heterogeneous distributed
  systems.
\newblock {\em arXiv preprint arXiv:1603.04467}, 2016.

\bibitem{badrinarayanan2015segnet}
V.~Badrinarayanan, A.~Kendall, and R.~Cipolla.
\newblock Segnet: A deep convolutional encoder-decoder architecture for image
  segmentation.
\newblock {\em arXiv preprint arXiv:1511.00561}, 2015.

\bibitem{bearman2016s}
A.~Bearman, O.~Russakovsky, V.~Ferrari, and L.~Fei-Fei.
\newblock What’s the point: Semantic segmentation with point supervision.
\newblock In {\em ECCV}. Springer, 2016.

\bibitem{chen2015semantic}
L.-C. Chen, G.~Papandreou, I.~Kokkinos, K.~Murphy, and A.~L. Yuille.
\newblock Semantic image segmentation with deep convolutional nets and fully
  connected crfs.
\newblock In {\em ICLR}, 2015.

\bibitem{cordts2016cityscape}
M.~Cordts, M.~Omran, S.~Ramos, T.~Rehfeld, M.~Enzweiler, R.~Benenson,
  U.~Franke, S.~Roth, and B.~Schiele.
\newblock The cityscapes dataset for semantic urban scene understanding.
\newblock In {\em CVPR}. IEEE, 2016.

\bibitem{farabet2013learning}
C.~Farabet, C.~Couprie, L.~Najman, and Y.~LeCun.
\newblock Learning hierarchical features for scene labeling.
\newblock {\em IEEE transactions on pattern analysis and machine intelligence},
  35(8):1915--1929, 2013.

\bibitem{ganin2015unsupervised}
Y.~Ganin and V.~Lempitsky.
\newblock Unsupervised domain adaptation by backpropagation.
\newblock In {\em ICML}, 2015.

\bibitem{ganin2016domain}
Y.~Ganin, E.~Ustinova, H.~Ajakan, P.~Germain, H.~Larochelle, F.~Laviolette,
  M.~Marchand, and V.~Lempitsky.
\newblock Domain-adversarial training of neural networks.
\newblock {\em Journal of Machine Learning Research}, 17(59):1--35, 2016.

\bibitem{goodfellow2014generative}
I.~Goodfellow, J.~Pouget-Abadie, M.~Mirza, B.~Xu, D.~Warde-Farley, S.~Ozair,
  A.~Courville, and Y.~Bengio.
\newblock Generative adversarial nets.
\newblock In {\em NIPS}, 2014.

\bibitem{guillaumin2014imagenet}
M.~Guillaumin, D.~K{\"u}ttel, and V.~Ferrari.
\newblock Imagenet auto-annotation with segmentation propagation.
\newblock In {\em IJCV}. Springer, 2014.

\bibitem{hoffman2016fcns}
J.~Hoffman, D.~Wang, F.~Yu, and T.~Darrell.
\newblock {FCN}s in the wild: Pixel-level adversarial and constraint-based
  adaptation.
\newblock {\em arXiv preprint arXiv:1612.02649}, 2016.

\bibitem{khosla2012undoing}
A.~Khosla, T.~Zhou, T.~Malisiewicz, A.~A. Efros, and A.~Torralba.
\newblock Undoing the damage of dataset bias.
\newblock In {\em ECCV}. Springer, 2012.

\bibitem{kingma2015adam}
D.~Kingma and J.~Ba.
\newblock Adam: A method for stochastic optimization.
\newblock In {\em ICLR}, 2015.

\bibitem{kolesnikov2016seed}
A.~Kolesnikov and C.~H. Lampert.
\newblock Seed, expand and constrain: Three principles for weakly-supervised
  image segmentation.
\newblock In {\em ECCV}. Springer, 2016.

\bibitem{lin2016scribblesup}
D.~Lin, J.~Dai, J.~Jia, K.~He, and J.~Sun.
\newblock Scribblesup: Scribble-supervised convolutional networks for semantic
  segmentation.
\newblock In {\em CVPR}. IEEE, 2016.

\bibitem{liu2016coupled}
M.-Y. Liu and O.~Tuzel.
\newblock Coupled generative adversarial networks.
\newblock In {\em NIPS}, 2016.

\bibitem{liu2011entropy}
M.-Y. Liu, O.~Tuzel, S.~Ramalingam, and R.~Chellappa.
\newblock Entropy rate superpixel segmentation.
\newblock In {\em CVPR}. IEEE, 2011.

\bibitem{long2015fully}
J.~Long, E.~Shelhamer, and T.~Darrell.
\newblock Fully convolutional networks for semantic segmentation.
\newblock In {\em CVPR}, 2015.

\bibitem{long2015learning}
M.~Long, Y.~Cao, J.~Wang, and M.~I. Jordan.
\newblock Learning transferable features with deep adaptation networks.
\newblock In {\em ICML}, 2015.

\bibitem{long2013transfer}
M.~Long, J.~Wang, G.~Ding, J.~Sun, and P.~S. Yu.
\newblock Transfer feature learning with joint distribution adaptation.
\newblock In {\em ICCV}, 2013.

\bibitem{long2016unsupervised}
M.~Long, H.~Zhu, J.~Wang, and M.~I. Jordan.
\newblock Unsupervised domain adaptation with residual transfer networks.
\newblock In {\em NIPS}, 2016.

\bibitem{noh2015learning}
H.~Noh, S.~Hong, and B.~Han.
\newblock Learning deconvolution network for semantic segmentation.
\newblock In {\em ICCV}. IEEE, 2015.

\bibitem{pan2010survey}
S.~J. Pan and Q.~Yang.
\newblock A survey on transfer learning.
\newblock {\em IEEE Transactions on knowledge and data engineering},
  22(10):1345--1359, 2010.

\bibitem{papandreou2015weakly}
G.~Papandreou, L.-C. Chen, K.~P. Murphy, and A.~L. Yuille.
\newblock Weakly-and semi-supervised learning of a deep convolutional network
  for semantic image segmentation.
\newblock In {\em ICCV}. IEEE, 2015.

\bibitem{pathak2015constrained}
D.~Pathak, P.~Krahenbuhl, and T.~Darrell.
\newblock Constrained convolutional neural networks for weakly supervised
  segmentation.
\newblock In {\em ICCV}. IEEE, 2015.

\bibitem{pathak2015fully}
D.~Pathak, E.~Shelhamer, J.~Long, and T.~Darrell.
\newblock Fully convolutional multi-class multiple instance learning.
\newblock 2015.

\bibitem{pinheiro2015image}
P.~O. Pinheiro and R.~Collobert.
\newblock From image-level to pixel-level labeling with convolutional networks.
\newblock In {\em CVPR}. IEEE, 2015.

\bibitem{purushotham2017variational}
S.~Purushotham, W.~Carvalho, T.~Nilanon, and Y.~Liu.
\newblock Variational recurrent adversarial deep domain adaptation.
\newblock In {\em ICLR}, 2017.

\bibitem{radford2015unsupervised}
A.~Radford, L.~Metz, and S.~Chintala.
\newblock Unsupervised representation learning with deep convolutional
  generative adversarial networks.
\newblock In {\em ICLR}, 2016.

\bibitem{richter2016playing}
S.~R. Richter, V.~Vineet, S.~Roth, and V.~Koltun.
\newblock Playing for data: Ground truth from computer games.
\newblock In {\em ECCV}. Springer, 2016.

\bibitem{ros2016synthia}
G.~Ros, L.~Sellart, J.~Materzynska, D.~Vazquez, and A.~M. Lopez.
\newblock The synthia dataset: A large collection of synthetic images for
  semantic segmentation of urban scenes.
\newblock In {\em CVPR}. IEEE, 2016.

\bibitem{saleh2016built}
F.~Saleh, M.~S.~A. Akbarian, M.~Salzmann, L.~Petersson, S.~Gould, and J.~M.
  Alvarez.
\newblock Built-in foreground/background prior for weakly-supervised semantic
  segmentation.
\newblock In {\em ECCV}. Springer, 2016.

\bibitem{sener2016learning}
O.~Sener, H.~O. Song, A.~Saxena, and S.~Savarese.
\newblock Learning transferrable representations for unsupervised domain
  adaptation.
\newblock In {\em NIPS}, 2016.

\bibitem{shimoda2016distinct}
W.~Shimoda and K.~Yanai.
\newblock Distinct class-specific saliency maps for weakly supervised semantic
  segmentation.
\newblock In {\em ECCV}. Springer, 2016.

\bibitem{torralba2011unbiased}
A.~Torralba and A.~A. Efros.
\newblock Unbiased look at dataset bias.
\newblock In {\em CVPR}. IEEE, 2011.

\bibitem{weinzaepfel2013deepflow}
P.~Weinzaepfel, J.~Revaud, Z.~Harchaoui, and C.~Schmid.
\newblock Deepflow: Large displacement optical flow with deep matching.
\newblock In {\em ICCV}. IEEE, 2013.

\bibitem{xie2016semantic}
J.~Xie, M.~Kiefel, M.-T. Sun, and A.~Geiger.
\newblock Semantic instance annotation of street scenes by 3d to 2d label
  transfer.
\newblock In {\em CVPR}. IEEE, 2016.

\bibitem{xu2015learning}
J.~Xu, A.~G. Schwing, and R.~Urtasun.
\newblock Learning to segment under various forms of weak supervision.
\newblock In {\em CVPR}. IEEE, 2015.

\bibitem{yu2016multi}
F.~Yu and V.~Koltun.
\newblock Multi-scale context aggregation by dilated convolutions.
\newblock In {\em ICLR}, 2016.

\bibitem{zellinger2017central}
W.~Zellinger, T.~Grubinger, E.~Lughofer, T.~Natschl{\"a}ger, and
  S.~Saminger-Platz.
\newblock Central moment discrepancy ({CMD}) for domain-invariant
  representation learning.
\newblock In {\em ICLR}, 2017.

\bibitem{zhu2016generative}
J.-Y. Zhu, P.~Kr{\"a}henb{\"u}hl, E.~Shechtman, and A.~A. Efros.
\newblock Generative visual manipulation on the natural image manifold.
\newblock In {\em ECCV}. Springer, 2016.

\end{thebibliography}

}
\clearpage
\appendix
\noindent\textbf{\Large Appendix}
\section{Visualize GA, CA and Static-Object prior}\label{Appendix_1}
In Sec.~\ref{GA}-\ref{prior} of the main paper, we explain how each component in our structure enhance the performance of segmentation, and also show quantitative results in experiment. Here we'll further illustrate effects of these components:

\begin{figure}[b]
	\centering
    \vspace{-5mm}
	\includegraphics[width=0.45\textwidth]{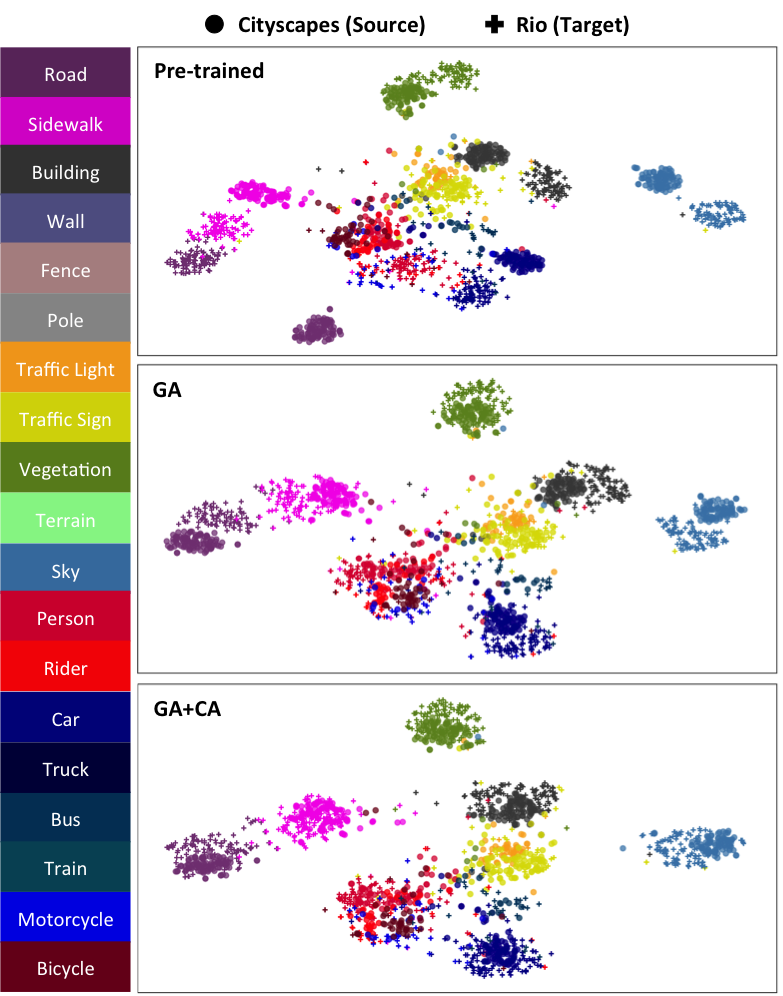}    
	\caption{\small t-SNE visualization results. For simplicity, we only show the results of the task \textit{Cityscapes $\rightarrow$ Rio}. We could clearly observe that the alignment between domains becomes better from \textit{pre-trained} to \textit{GA+CA}.}
    \vspace{-5mm}
	\label{fig.t-SNE}
\end{figure}

\noindent\textbf{T-SNE Visualization}
To visualize the adaptation results on common feature space with t-SNE, we randomly select 100 images from each domain, and for each image we extracted its average $fc7$ feature from each class, so for both source and target we have 100 feature points from each class. 

As shown in Fig.~\ref{fig.t-SNE}, with pre-trained model only, there is an obvious shift between source and target domain. After applying the global alignment (GA), the distance between clusters with same labels becomes closer, while we could still observe a gap between domains. Once we further apply the class-wise alignment (CA), the gap between domains nearly vanishes. This result again demonstrates the effectiveness of each component of our proposed method. 

\begin{figure}[h]
\begin{center}
	\includegraphics[width=0.45\textwidth]
{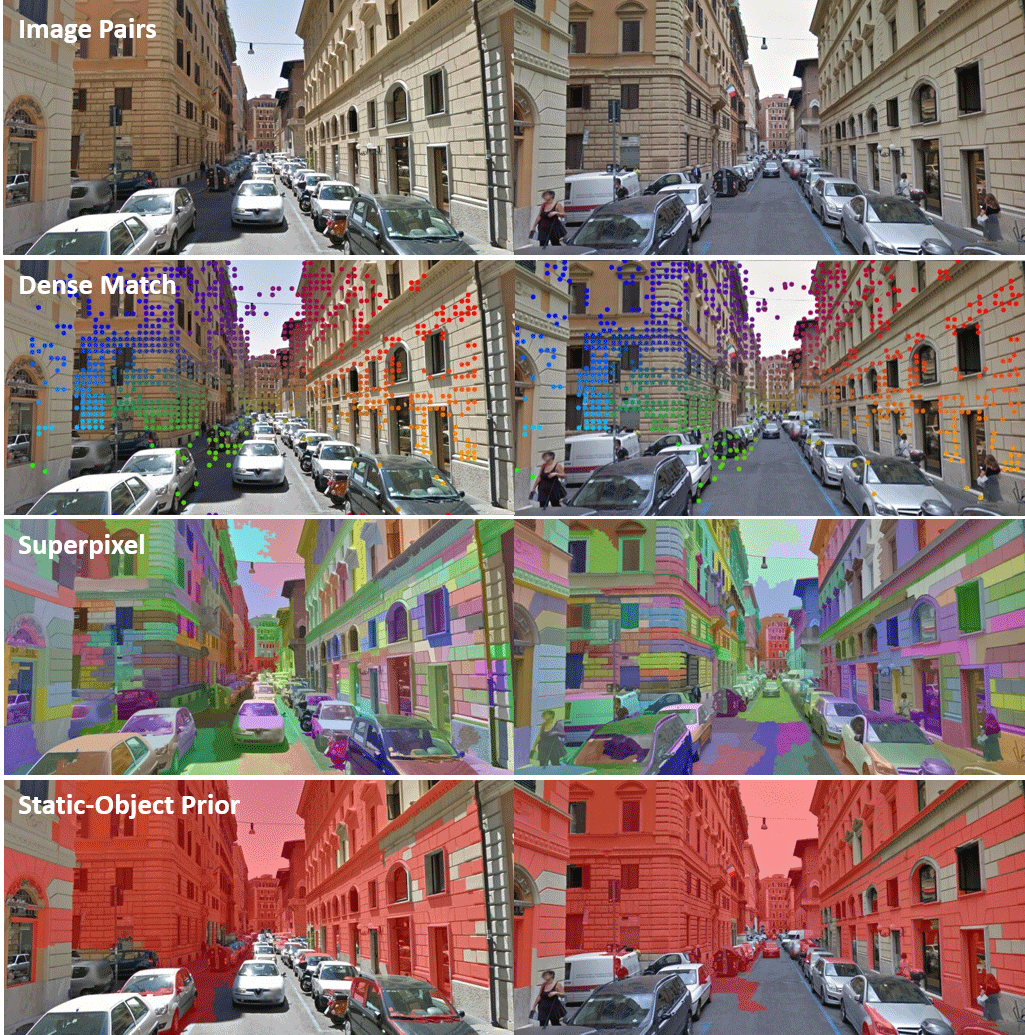}
    \caption{\small Typical results of our static-object prior pipeline. The first row is the original unlabeled image pair of same place across time. The second row is the result of dense matching, noted by points of same color. The third row is the result of superpixel segmentation marked by different colors. Combining the results from the above two rows, we could extract static-object prior of this image pair, as shown by the red regions in the last row. }
	\vspace{-5mm}
	\label{fig.sp_example}
    \end{center}
\end{figure}

\noindent\textbf{Harvesting Static-Object Prior} In Sec.~\ref{prior}, we propose a novel pipeline to extract the static-object prior using the natural synchronization of static objects over time. For better understanding, we show some typical results of our proposed pipeline in Fig.~\ref{fig.sp_example}. Clearly, most of the regions identified by our method truly belong to static-objects. This demonstrates the effectiveness of our method.

\begin{figure*}[t]
      \begin{center}
      \includegraphics[width=0.9\textwidth]{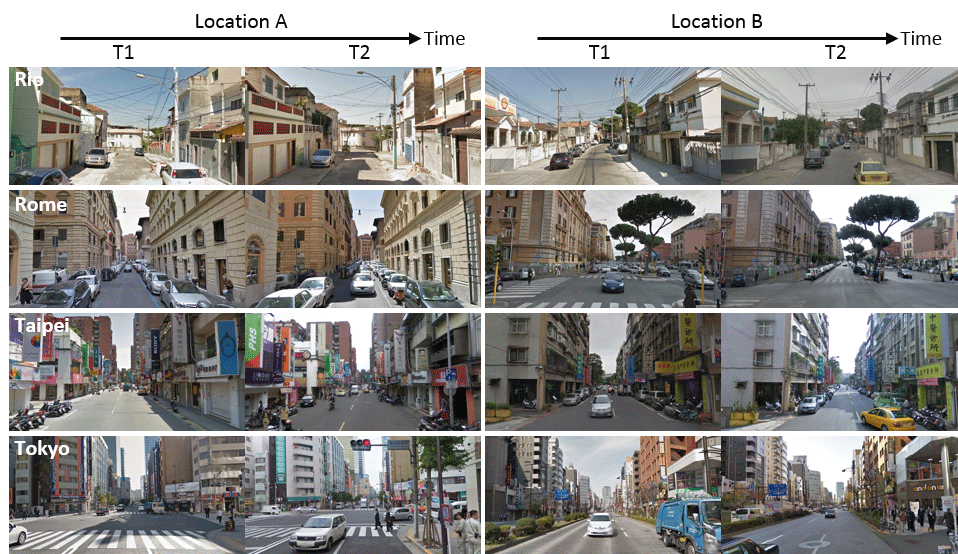}
      \end{center}
      \caption{\small Examples of the unlabeled image pairs of different cities in our dataset. In each row, we show two image pairs at different locations in one city.}	
      \label{fig.time_machine}    
 \end{figure*}
 
\begin{figure*}[!t]
	\begin{center}
	\includegraphics[width=0.9\textwidth]{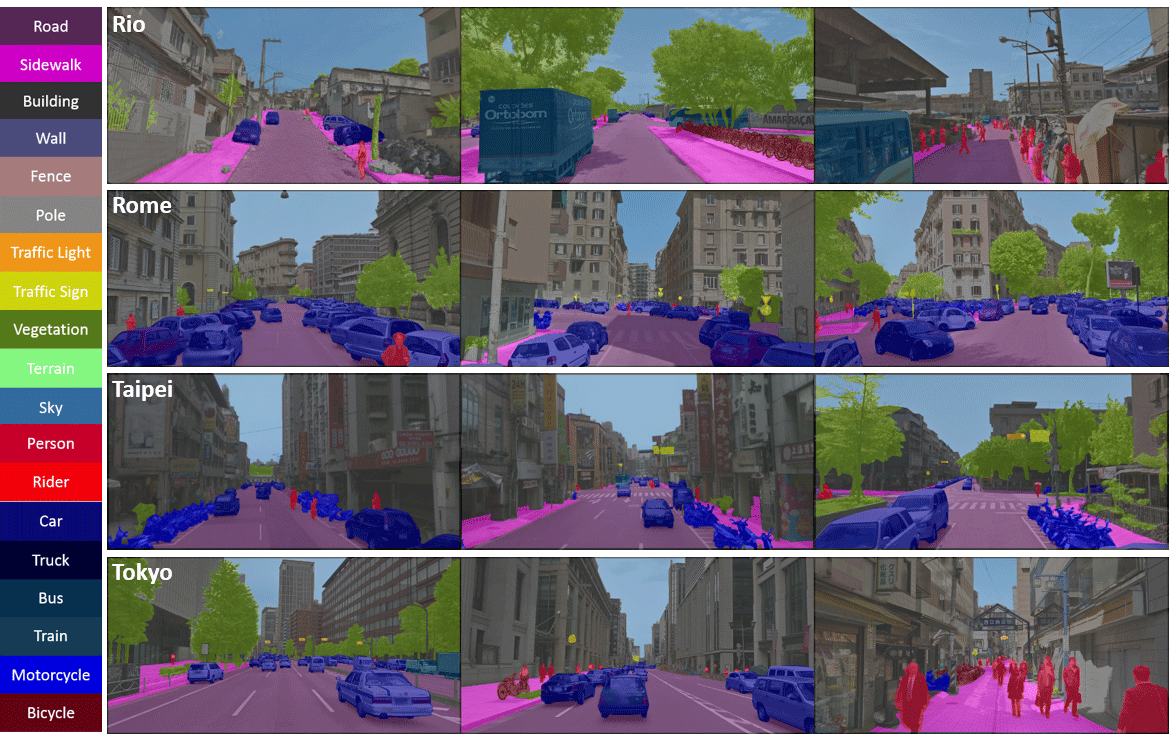}
	\end{center}
	\vspace{-3mm}
	\caption{\small Examples of the labeled images of different cities in our dataset. Each image is annotated in good quality.}
	\label{fig.annotation}
\end{figure*}

\section{Dataset}\label{Appendix_2}
To demonstrate the uniqueness of our dataset for road scene semantic segmenter adaptation, here we show more examples of it. 

\noindent\textbf{Unlabeled Image Pairs} There are more examples collected at different cities with diverse appearances in Fig.~\ref{fig.time_machine}. Valuable temporal information which facilitates \textit{unsupervised} adaptation is contained in these image pairs. 

\noindent\textbf{Labeled Image} We also show more annotated images in Fig.~\ref{fig.annotation} to demonstrate the label-quality of our dataset.  


\section{ Synthetic to Real Adaptation}\label{Appendix_3}
In Sec.~5.4 of the main paper, we have shown the quantitative results of this adaptation task in Table~3. We conclude that our method could perform well even under this challenging setting. To better support our conclusion, here we show some typical examples of this task in Fig.~\ref{fig.synthetic}.

\begin{center}
    \centering
    \includegraphics[trim=0.0in 0.0in 0.0in 0in,width=\linewidth]{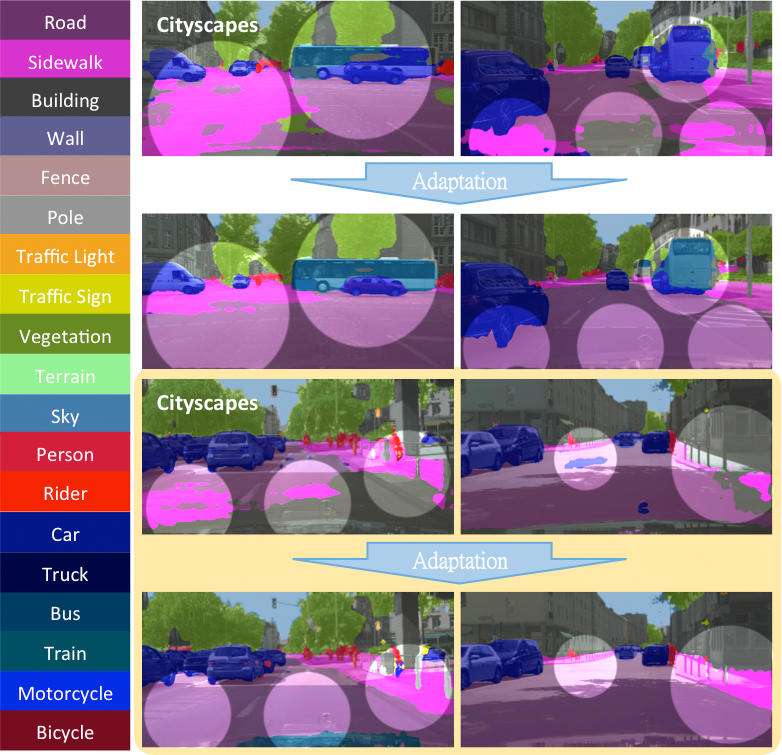}
    \captionof{figure}{\small
adaptation task: STNTHIA to Cityscapes. The first row and second show the results before and after adaptation, respectively.
}
    \label{fig.synthetic}
\end{center}

\end{document}